\documentclass{article}

\usepackage{microtype}
\usepackage{graphicx}
\usepackage{subfigure}
\usepackage{multirow}
\usepackage{gensymb}
\usepackage{amssymb}
\usepackage{float}

\usepackage{booktabs} 

\usepackage{hyperref}

\usepackage[accepted]{icml2018}

\icmltitlerunning{Sample Efficient Semantic Segmentation  using Rotation Equivariant Convolutional Networks}

\begin{document}

\twocolumn[
\icmltitle{Sample Efficient Semantic Segmentation  using\\ Rotation Equivariant Convolutional Networks}



\icmlsetsymbol{equal}{*}

\begin{icmlauthorlist}
\icmlauthor{Jasper Linmans}{equal,uva}
\icmlauthor{Jim Winkens}{equal,uva}
\icmlauthor{Bastiaan S. Veeling}{uva}
\icmlauthor{Taco S. Cohen}{uva}
\icmlauthor{Max Welling}{uva}
\end{icmlauthorlist}

\icmlaffiliation{uva}{University of Amsterdam}

\icmlcorrespondingauthor{Jasper Linmans}{jasperlinmans@gmail.com}
\icmlcorrespondingauthor{Jim Winkens}{jimwinkens@gmail.com}

\icmlkeywords{Machine Learning, ICML}

\vskip 0.3in]



\printAffiliationsAndNotice{\icmlEqualContribution} 

\begin{abstract}
We propose a semantic segmentation model that exploits rotation and reflection symmetries. We demonstrate significant gains in sample efficiency due to increased weight sharing, as well as improvements in robustness to symmetry transformations. The group equivariant CNN framework is extended for segmentation by introducing a new equivariant $(G\rightarrow\mathbb{Z}^2)$-convolution that transforms feature maps on a group to planar feature maps. Also, equivariant transposed convolution is formulated for up-sampling in an encoder-decoder network. To demonstrate improvements in sample efficiency we evaluate on multiple data regimes of a  rotation-equivariant segmentation task: cancer metastases detection in histopathology images. We further show the effectiveness of exploiting more symmetries by varying the size of the group.
\end{abstract}

\section{Introduction}

Convolutional networks are able to model local patterns in data efficiently by sharing parameters in the convolutional layers. In doing so, the translational symmetry of images is preserved: shifting a layer's input produces a proportionate shift $T$ in the layer's output, i.e. $f(T\mathbf{x})=Tf(\mathbf{x})$. This property known as translation equivariance exploits the spatial structure inherent to many sensory data like images, video and audio.

It has been shown that the effectiveness of exploiting symmetries extends beyond mere translations. Recent work by \cite{Cohen2016-do} on group equivariant convolutions has explored a higher degree of weight sharing to encode discrete rotation and reflection symmetries.
The resulting rotation equivariant CNN exhibits an increased expressive power and in turn a higher sample efficiency, without an increase in the number of parameters. These networks have shown considerable gains in classification performance, but have not been extended to function in a segmentation setting.

\begin{figure*}[ht]
    \centering
    \includegraphics[width=\textwidth]{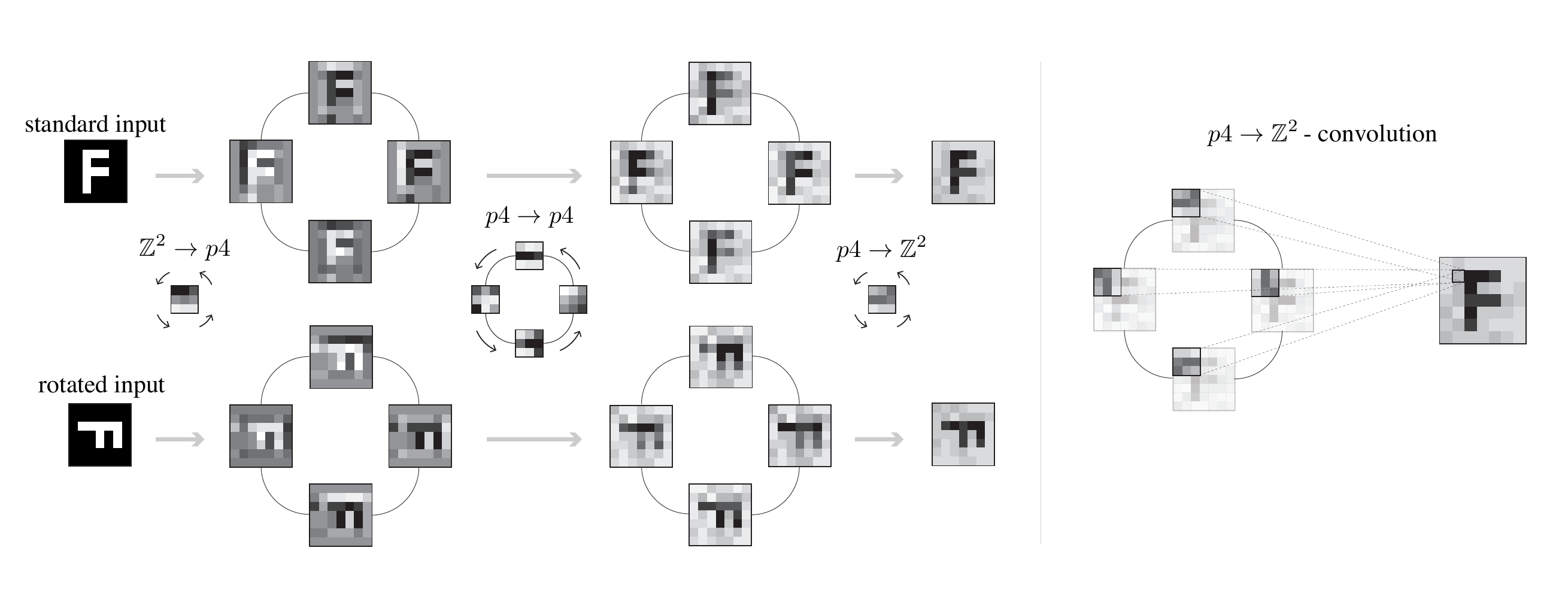}
    
    \caption{Given a canonical input and a rotated duplicate, we demonstrate how a 2-layer G-CNN is equivariant in $p4$.  The $\mathbb{Z}^2\rightarrow p4$ convolution correlates the input with 4 rotated versions of the same kernel. The $p4\rightarrow p4$ convolution correlates the resulting feature map with the $p4$-kernel, cyclically-shifting and rotating the kernel for each orientation. The final $p4\rightarrow \mathbb{Z}^2$ layer correlates the $p4$ feature map with a single planar filter according to the transformations defined by the group. Note that rotating and cyclically shifting the intermediate $p4$ feature maps of the standard input indeed produces those of the rotated version.}
    \label{equiv}
    \vspace{-.1cm}
\end{figure*}

We leverage the group convolutions of \cite{Cohen2016-do} to encode rotation and reflection symmetries in a pixel-wise segmentation model. To enable segmentation for this framework, we extend it with a  convolution operation that maps from a group representation to the $\mathbb{Z}^2$ grid and we use group convolution in a transposed convolution layer, which we show to be equivariant. 

Domains that benefit most from exploiting rotation symmetries are those that lack a  canonical orientation, such as medical imaging and in particular histopathology data. This domain is especially well-suited for the use of equivariant convolutional networks; the limited availability of pixel-level annotations requires a high sample efficiency. As such, to demonstrate the potential of the proposed model we evaluate on a histopathology dataset derived from the Camelyon16 challenge \cite{bejnordi2017diagnostic}. We show that the increased weight sharing by explicitly encoding rotation and reflection symmetries leads to consistent performance gains, especially under limited training dataset size.

We further establish that conventional CNNs trained on histopathology data demonstrate erratic predictions under $\pi/2$ rotations and reflections. Such behavior is alleviated by the proposed group equivariant model which helps accommodate the requirements of model predictability in a clinical setting.

\subsection*{Related work}
A common approach to improve robustness to rotations and reflections for CNNs is to use extensive data augmentation during training \cite{Liu2017-jq, litjens2017survey}. Although this potentially leads to better generalization, it also increases the amount of training samples. Furthermore, the additional rotated copies of training data induce the model to learn permuted copies of the same filter, adding redundancy to the network's weights and increasing the risk of overfitting. Our proposed model reduces these effects with built-in equivariance, and hence could exhibit an improved sample efficiency. Additionally even if rotation equivariance is achieved on the training data, there is no guarantee that this generalizes to a test set. To approximate test time equivariance, \cite{Liu2017-jq, Ciresan2013-wv} propose a test-time augmentation strategy that averages the predictions of $90\degree$-rotated and mirrored versions. This however increases the computational cost of inference by a factor of eight and does not guarantee equivariance \cite{DBLP:journals/corr/LencV14}.

To achieve equivariance, we focus on the straight-forward G-CNN framework from \cite{Cohen2016-do} applied on discrete rotation and reflection groups. Work by \cite{Worrall2017-ji} further exploits rotational symmetries allowing for full $360\degree$-equivariance, such Harmonic  Networks constrain the set of filters  to  circular harmonics. \cite{Weiler2017-oz} employs steerable filters and uniformly samples a small number of rotations. Although the groups in the G-CNN framework from \cite{Cohen2016-do} do not cover the full continuous rotational symmetries, the empirical evidence gathered so far shows that $90\degree$ rotation equivariance improves performance significantly \cite{Weiler2017-oz, veeling2018}. Similar work by \cite{veeling2018} leverages the G-CNN framework on histopathology data in a classification setting, demonstrating consistent performance gains and improvements in sample efficiency when comparing an equivariant model with an equivalent standard CNN.

\section{Methodology}

\begin{figure*}[ht]
    \centering
    
    \includegraphics[width=\textwidth]{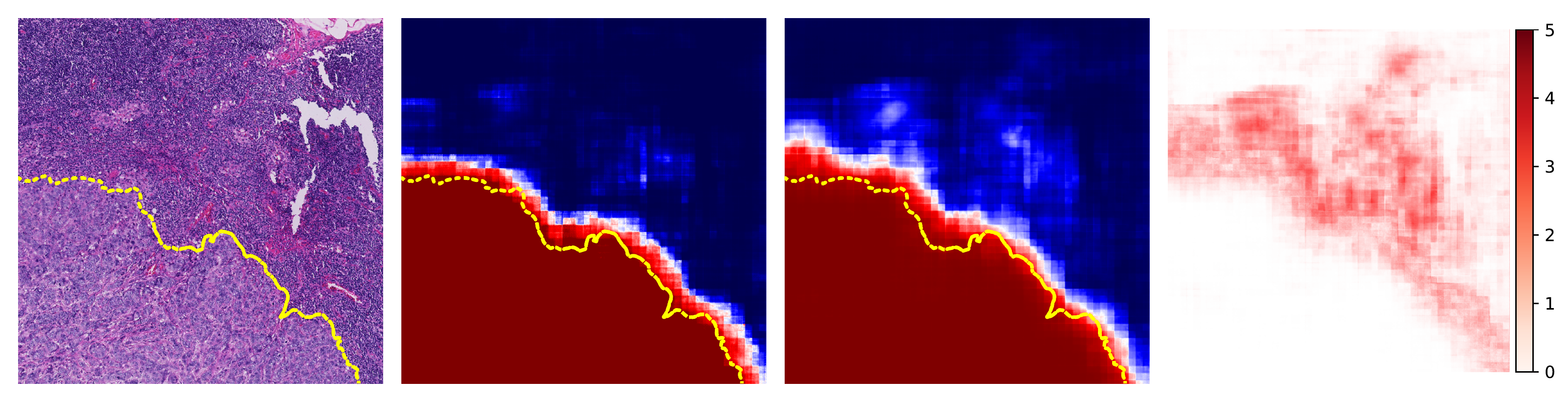}
    
    \caption{\textbf{(a)} Example of a $630\mu m^2$ patch and its ground truth tumor mask (marked by the yellow border), \textbf{(b)} the mean tumor prediction of the patch across the eight roto-reflection transformations in $p4m$ as viewed in the original orientation using the P4M U-Net and \textbf{(c)} using the baseline U-Net. \textbf{(d)}  Standard deviation of the tumor predictions across the roto-reflection transformations using the baseline U-Net. The standard deviation of the P4M U-Net model is omitted since the values are not visible in the same range and correspond to the accumulated numerical error.}
    \label{model_stability}
    \vspace{-.1cm}
\end{figure*}
\subsection{Mathematical framework}


Here, we further develop the mathematical framework of G-CNNs \cite{Cohen2016-do} such that it can be used in a segmentation setting, by formalizing transposed group convolutions and introducing a new convolutional operation transforming group feature maps to planar feature maps. The implementation of the equivariant layers is available on: \texttt{https://github.com/nom/gcnn-seg}. 

G-CNNs utilize group convolutions, which enjoy increased weight sharing and better statistical efficiency than regular convolutions. Specifically, it is implemented for the $p4$ group, consisting of translations and rotations by multiples of $\pi/2$, and the $p4m$ group, which additionally includes reflections. G-CNNs are generalizations of CNNs where feature maps are considered as functions on these groups, e.g. for the $p4m$ group each feature map contains eight orientation channels, corresponding to the number of roto-reflections in the group. In Fig. \ref{equiv} we visualize such G-CNN feature maps and demonstrate how equivariance is preserved throughout the different group convolutions.

The first layer transforms an input image $f : \mathbb{Z}^2 \rightarrow \mathbb{R}^K$, with $K$ the number of channels using a filter $\psi$, which is defined as the ($\mathbb{Z}^2 \rightarrow G$)-convolution:
\begin{equation}
    [f * \psi](g) = \sum_{y \in \mathbb{Z}^2} \sum_{k=1}^K f_k(y) \psi_k(g^{-1} y),
\end{equation}
where $g = (r, t)$ is a roto-translation (in case $G=p4$) or a roto-reflection-translation (in case $G=p4m$).

In the next layers, the feature maps are functions on $G$, hence the filters must also be functions on $G$, for which  the $(G\rightarrow G)$-convolution is used:
\begin{equation}
    [f * \psi](g) = \sum_{h\in G}\sum^K_{k=1} f_k(h)\psi_k(g^{-1} h).
\end{equation}
The transpose of this linear operation, the \textit{transposed} convolution, is used to increase the spatial size of the feature maps. Contrary to strided group convolution \cite{Cohen2016-do}, the choice of stride for transposed group convolution does not affect equivariance.


To allow for the equivariant transformation of a feature map in $G$ to a two-dimensional segmentation mask  $m : \mathbb{Z}^2 \rightarrow \mathbb{R}^C$, with $C$ the number of classes, we propose the definition of a $(G \rightarrow \mathbb{Z}^2 )$-convolution:
\begin{equation}
    [f * \psi](y) = \sum_{h\in G}\sum^K_{k=1} f_k(h)\psi_k(z(y)^{-1}h),
    \label{eq:gtoz2}
    \end{equation}
where $z(y)$ is the translation by $y$ in $\mathbb{Z}^2$.

Contrary to the $(G\rightarrow G)$-convolution, the filter $\psi$ in Eq. \ref{eq:gtoz2} is  a function on $\mathbb{Z}^2$. This single planar filter is shared across the orientation channels following the transformations of the group,  similar to the ($\mathbb{Z}^2 \rightarrow G$)-convolution (see Fig. \ref{equiv}). This introduces a learnable transformation from a feature map in $G$ to a segmentation mask which is naturally more expressive then pooling the orientation channels \cite{Cohen2016-do, Weiler2017-oz}.

\subsection{GU-Net architecture}
To obtain pixel-wise segmentation maps, we use the conventional U-Net architecture \cite{ronneberger2015u} as a baseline for our rotation equivariant model. The \textit{GU-Net} architecture is constructed by replacing all (transposed) convolution layers with their group equivariant counterparts. 
Two-layer convolution blocks are followed by a $3 \times 3$ max-pool that incrementally reduces the spatial size, up to a depth level of four. Then pooling is replaced by $3 \times 3$ transposed convolutions with zero padding to recover the spatial size of the input image and enable per-pixel classification. Batch normalization is applied after every convolution operation, including transposed convolutions. For the \textit{GU-Net} version, these batch  normalization  operations  are  made group equivariant by aggregating moments per group feature map rather than spatial  feature  map. The proposed $(G \rightarrow \mathbb{Z}^2 )$-convolution is finally used to transform the orientation channels to the two-dimensional grid of the output mask.

To ensure a fair comparison between the proposed equivariant model and the baseline model, the number of parameters is kept constant \cite{Cohen2016-do}. To this end the amount of filters used by the G-CNN model is divided by the square root of the group size. This way the additional parameters introduced by an increase in orientation channels, increasing each layers input size, are compensated for.

\section{Experiments}


\subsection{Dataset}
The proposed model is evaluated on the PatchCamelyon dataset \cite{veeling2018} derived from the Camelyon16 challenge \cite{bejnordi2017diagnostic} with the task of tumor localization. The original challenge data contains 400 H\&E stained whole slide images (WSIs) of sentinel lymph node sections with pixel-level annotations. The PatchCamelyon dataset consists of 300.000 patches of $320\times320$ pixels at 10$\times$ magnification, with a 8:1:1 split into training, validation and test set. To prevent selecting background and non-tissue patches, patches are converted to HSV, blurred and selected if max pixel saturation lies above 0.07 (range $[0,1]$), and value above 0.1. This was empirically verified to not drop tissue patches. The patches were extracted by uniformly sampling WSIs and drawing tumor/non-tumor patches with equal probability. 
\subsection{Training details} Models are optimized using Adam \cite{KingmaB14} with initial learning rate $1\mathrm{e}{-3}$ (halved after 20 epochs of no improvement in validation loss). Epochs consist of 468 batches with a batch size of 64. Weights with lowest validation loss are selected for test evaluation.



%

\subsection{Model stability}
To assess the predictive stability of the P4M U-Net model, we examine the standard deviation of predictive probabilities  under roto-reflection transformations of the input as compared to the baseline model. 
 Fig. \ref{model_stability} shows the analysis for an example patch. 
  We observe that the standard U-Net is prone to erratic prediction behavior under roto-reflections, especially for uncommon patterns that occur only in a limited set of orientations in the training data such as tumor boundaries.


\subsection{Segmentation performance}
The segmentation performance of the proposed model is evaluated using the Dice similarity coefficient (DSC), Table \ref{data-regimes} reports the results. To compare with a data augmentation strategy, the training data is augmented with roto-reflection transformations for the baseline U-Net. To compare the sample efficiency, we experimented with multiple data regimes, where the number of samples in the training set is incrementally reduced by a factor of two. 

\begin{table}[H]

\centering
\setlength{\tabcolsep}{10pt} 
\caption{Performance on the Camelyon16 derived dataset.}
 \begin{tabular}{@{}llc}
\toprule
Model & Data  & DSC \\ \midrule
\multirow{4}{*}{\shortstack[l]{P4M U-Net}} & $100\%$ & 83.7       \\
& $50\%$ & 82.1 \\
& $25\%$ & 81.6 \\
& $12.5\%$ & 79.6\\ 
\midrule
\multirow{4}{*}{\shortstack[l]{U-Net}} & $100\%$ & 79.8\\
& $50\%$ & 78.5 \\
& $25\%$ & 77.6 \\
& $12.5\%$ & 74.2\\ 
\bottomrule
\end{tabular}
\label{data-regimes}
\end{table}

These results show that the proposed model performs consistently better than the baseline method in terms of the DSC metric, and when trained on the 25\% data regime it even outperforms the baseline model trained on the entire dataset. We see that the superiority of our proposed architecture is predominantly due to the increased weight sharing in the $p4m$-equivariant model, which frees up model capacity and reduces the redundancy of detecting the same local patterns in different orientations.

We also observe that the performance gap between our model and the baseline increases when we limit the dataset size by removing training samples. For example, our method outperforms the baseline on 1/8th of the data by 7.2\%. This seems to indicate that the performance in the small-data regime benefits significantly from the sample efficiency of P4M U-Net, with diminishing returns when the amount of data is sufficient for the baseline network to achieve (approximate) rotation equivariance. This performance gap remains for the full data set.

We further vary the group size to only roto-transformations to study whether adding symmetries contributes to a better performance. For the $p4$ group, we measure a Dice coefficient on the full test set of 82.5 compared to 83.7 for the $p4m$ group, which indicates that additionally exploiting the reflection symmetry corresponds to better generalization.





\section{Conclusion}
By extending the group equivariant CNN framework and using it in an encoder-decoder architecture for pixel-wise predictions, we show a consistent performance increase compared to an equivalent conventional CNN. We further demonstrate that the performance gap between the G-CNN and the baseline model is most significant for smaller data regimes, highlighting a substantial enhancement of sample efficiency by G-CNNs. We also confirm experimentally, by increasing the size of the group, that exploiting more symmetries leads to better generalization. Notably, we further demonstrate that conventional CNNs show erratic predictions for rotations and reflections which is alleviated by the use of group convolutions.







\end{document}